\documentclass[conference]{IEEEtran}
\IEEEoverridecommandlockouts
\usepackage{cite}
\usepackage{amsmath,amssymb,amsfonts}
\usepackage{algorithmic}
\usepackage{graphicx}
\usepackage{tabularx}
\usepackage{textcomp}
\usepackage{xcolor}
\usepackage{url}
\usepackage{multirow} % preamble에 추가 필요
\usepackage{booktabs} 
\def\BibTeX{{\rm B\kern-.05em{\sc i\kern-.025em b}\kern-.08em
    T\kern-.1667em\lower.7ex\hbox{E}\kern-.125emX}}
\begin{document}

\title{Analysing the Language of Neural Audio Codecs\\
}

\author{\IEEEauthorblockN{Joonyong Park$^{1}$, Shinnosuke Takamichi$^{1,2}$, David M. Chan$^{3}$, Shunsuke Kando$^{1}$, Yuki Saito$^{1}$, Hiroshi Saruwatari$^{1}$}
\IEEEauthorblockA{$^1$The University of Tokyo, $^2$Keio University, $^3$University of California, Berkeley. 
\\joonyong-park@g.ecc.u-tokyo.ac.jp}
}

\maketitle

\begin{abstract}

This study presents a comparative analysis of the statistical and linguistic properties of neural audio codecs (NACs). We investigate discrete speech tokens produced by various NAC models, examining their adherence to linguistic statistical laws such as Zipf's law and Heaps' law, as well as their entropy and redundancy. To assess how these token-level properties relate to semantic and acoustic preservation in synthesized speech, we evaluate intelligibility using error rates of automatic speech recognition, and quality using the UTMOS score. Our results reveal that NAC tokens, particularly 3-grams, exhibit language-like statistical patterns. Moreover, these properties, together with measures of information content, are found to correlate with improved performances in speech recognition and resynthesis tasks. These findings offer insights into the structure of NAC token sequences and inform the design of more effective generative speech models.

\end{abstract}

\vspace{-2mm}
\section{Introduction}

Speech discretisation allows models to take advantage of powerful sequence modelling techniques originally developed in natural language processing (NLP). Among such methods, neural audio codec (NAC) models have recently emerged as highly effective tools for modelling speech. They are capable of producing fine-grained token sequences capturing acoustic details essential for tasks such as automatic speech recognition (ASR), speech synthesis, and spoken language understanding~\cite{guo2025recent, borsos23}.

While NAC models have primarily been designed for efficient waveform compression and high-fidelity audio reconstruction, they are increasingly being integrated into generative and representation learning frameworks. Specifically, a speech token sequence derived from a pretrained NAC model serves as intermediate representations for downstream speech processing tasks. 
However, the linguistic and statistical properties of NAC-derived tokens remain underexplored. Determining whether these tokens exhibit language-like regularities is crucial, as it informs whether NLP-inspired or fundamentally new approaches are needed for generative speech modeling.

To address this open question, we conduct an in-depth analysis of NAC token sequences, comparing their statistical properties with those of natural languages. We focus on key linguistic regularities—including Zipf's law, Heaps' law, and entropy-based redundancy—to elucidate the extent to which NAC tokens adhere to, or deviate from, language-like statistical behaviours. 
Our analysis shows that these NAC-derived token sequences, particularly at the 3-gram level, closely follow natural language statistical laws, and this linguistic regularity correlates with the improved objective evaluation metrics of resynthesised speech.

\begin{figure}
    \centering
    \includegraphics[width=0.98\linewidth]{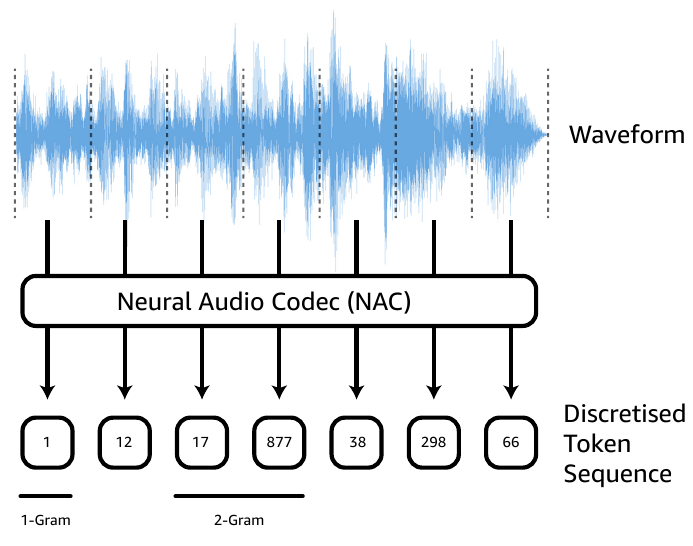}
    \caption{Analysis of NAC tokens conducted in our paper. An NAC model segments input audio into short frames and assigns each a symbolic token from a learned codebook, forming ``languages'' of audio tokens.}
    \label{fig:audio}
    \vspace{-5mm}
\end{figure}

\vspace{-2mm}
\section{Related Work}
\vspace{-1mm}
\subsection{Discrete Tokens from Speech}
Initially, self-supervised learning (SSL) models played a pivotal role in advancing speech tokenization. These models learn to produce discrete tokens directly from raw speech without relying on explicit phoneme labels or text transcriptions, by utilizing the discretisation methods such as $k$-means clustering. HuBERT~\cite{hsu2021hubert} and wav2vec 2.0~\cite{baevski2020wav2vec2} are well-known examples of these SSL models, whose token sequences have been shown to capture both phonetic and semantic information, thereby enhancing performance across a variety of speech processing tasks~\cite{Mohamed_2022}.

Building upon these foundations, NAC models, such as EnCodec~\cite{defossez2023high} and SoundStream~\cite{Zeghidour2021SoundStreamAE}, have further constructed the process with the explicit goal of speech resynthesis. NAC models were initially developed to compress audio for efficient data transmission by lowering the bitrate, encoding detailed acoustic information into compact token sequences. These models are optimized for efficient compression and high-fidelity waveform reconstruction, demonstrating high effectiveness in processing the audio. This emphasis on preserving fine-grained acoustic detail distinguishes NACs from SSL-based approaches and makes them particularly well-suited for generative speech tasks that require high-quality audio output.

Moreover, NAC models have been successfully applied in a variety of downstream tasks. For example, in text-to-speech (TTS) synthesis, NAC tokens enable high-quality waveform generation from text-based inputs, contributing to improved audio fidelity and naturalness~\cite{defossez2023high, Zeghidour2021SoundStreamAE}. NACs have also been integrated into ASR pipelines, where compressed token representations facilitate efficient and accurate transcription while reducing computational costs~\cite{dhawan2024codec}. Furthermore, NAC models have been explored for speech separation tasks, operating directly in the compressed token space to achieve efficient source separation with lower computational requirements~\cite{yip2024towards}. 
\vspace{-2mm}
\subsection{Statistical Laws of Language}

In this study, we explore three primary statistical laws, models, and metrics inspired by NLP techniques.

\vspace{0.5em}\noindent\textbf{Zipf's Law:\:} In linguistics and information theory, it is well established that the frequency distribution of words or character $n$-grams in natural language tends to follow Zipf's law~\cite{zipf1949human}, characterised by a few high-frequency elements and many low-frequency elements. For example, the third most frequent word in an English document, ``and,'' appears approximately one-third as often as the most frequent word, ``the\footnote{https://www.cs.cmu.edu/˜cburch/words/top.html}.'' This relationship is given by the following power-law equation between the frequency rank $r$ of a word and its frequency $f(r)$:
\vspace{-2mm}
\begin{align}
f(r) = a \cdot r^{-\eta},
\end{align}
where $a>0$ is a scaling constant and $\eta>0$ represents the sharpness or scaling properties of the distribution. An ideal Zipfian distribution typically has $\eta \approx 1$. Such distributions have been often observed not only in natural languages but also in animal communication systems and large language models, making $\eta$ a key indicator of the balance between redundancy and information efficiency~\cite{mandelbrot1953contribution, gelbukh2001zipf}.

\vspace{0.5em}\noindent\textbf{Heaps' Law:\:}
To capture the properties of a natural language in terms of the uniqueness of vocabulary, analyses based on Heaps' law~\cite{heaps1978information} are also employed. This law describes a sublinear relationship between the number of words in document \( m \) and the vocabulary size \( V(m) \), expressed as:
\vspace{-1mm}
\begin{align}
V(m) = K \cdot m^{\beta} \quad (0 < \beta < 1).
\end{align}
Here, \( K > 0 \) is a scaling factor that indicates the initial rate of vocabulary growth, while \( \beta \) determines the rate at which new vocabulary continues to be introduced as the text grows. A higher value of \( K \) suggests that the text introduces a relatively large number of unique words early on, whereas a lower value of \( K \) implies a more conservative initial vocabulary expansion. For values of  \( \beta \) close to 1, the vocabulary grows almost linearly with the document size, implying a rich and diverse vocabulary set. For lower \( \beta \) values, the vocabulary growth is sublinear, reflecting a language that heavily reuses existing words and introduces new ones less frequently.

\vspace{0.5em}\noindent\textbf{Entropy and Redundancy:\:}
Entropy, proposed by Shannon~\cite{shannon1951prediction}, is widely used as a measure of the diversity and unpredictability of information within a sequence and serves as a fundamental metric in evaluating linguistic systems and coding efficiency. The entropy $H$ of a token sequence is defined as:
\vspace{-2mm}
\begin{align}
H &= -\sum_{i=1}^{V} p_i \log_2 p_i,
\end{align}
where $V$ is the vocabulary size and $p_i$ is the probability of occurrence of the $i$-th token. Using entropy-based Huffman coding, it is possible to quantify the bit reduction rate and redundancy of a sequence. The average code length $L$ achieved by Huffman coding can be evaluated with respect to the entropy as follows:
\vspace{-1mm}
\begin{align}
H \leq L < H + 1.
\end{align}
By computing the ratio between the actual average code length $L$ and the entropy $H$, the compression efficiency or redundancy $R$ can be quantified as:
\vspace{-1mm}
\begin{align}
R = \frac{L - H}{L}.
\end{align}
A smaller value of $R$ indicates more efficient coding, reflecting lower redundancy and higher compressibility of the token.
\vspace{-2mm}
\subsection{Statistical Linguistic Analysis of Speech Tokens}

There have been some previous attempts to analyse discrete speech tokens using this type of analysis. 
Takamichi et al.~\cite{takamichi2024learned} analysed whether speech tokens generated by SSL models follow Zipf's law, as observed in natural language text. Their investigation revealed that these tokens exhibit power-law behaviour, which suggests that speech tokens possess certain linguistic structural properties. 
Moreover, Sicherman et al.~\cite{Sicherman2023AnalysingDS} explored the interpretability and redundancy of SSL-based speech tokens and found a strong correlation between tokens and phonemes. They introduced a method for identifying redundancies that can degrade the performance of language models, and they proposed techniques to reduce such redundancies, leading to improvements in downstream tasks like speech synthesis and generative spoken language modelling. 
With respect to NAC-based tokens, Liu et al.~\cite{liu2024visual} identified the challenge of inconsistency—different token sequences may be produced from the same speech input due to variations in the tokenization process. To address this, they introduced robust instruction-tuning strategies to enhance the consistency of tokens across different contexts, thereby improving the reliability of systems. 

\vspace{-2mm}
\section{Experimental Setup}  

Despite the above analyses, the question of whether NAC tokens inherit linguistic properties, such as statistical regularities or syntactic structures, remains underexplored. We thus aim to fill this gap by systematically analysing NAC token sequences to investigate their adherence to known linguistic properties and to clarify the relationships between speech metrics and linguistic structure in speech tokenization.

%need to show dataset  
% and kbps represents the bitrate in unit bits per second.

\begin{table}[t!]
\centering
\fontsize{8}{10}\selectfont
\setlength\tabcolsep{2pt}
\caption{Codec configuration comparison. ``A--F'' represents different NAC models, where ``A'' is SpeechTokenizer~\cite{zhang2023speechtokenizer}, ``B$\sim$'' is AcademiCodec~\cite{yang2023hifi}, ``C'' is AudioDec~\cite{wu2023audiodec}, ``D'' is DAC~\cite{kumar2023high}, ``E$\sim$'' is EnCodec~\cite{defossez2023high}, and ``F$\sim$'' is FunCodec~\cite{du2023funcodec}.  $n_d$ represents the dimension size, SR represents the sample rate. The contents of this table are partially cited from~\cite{wu2024audiolanguagemodeling}.}
\label{tab:nac_models}
\begin{tabularx}{0.48\textwidth}{lcccccc}
    \toprule
    & \textbf{Codec configurations} & \textbf{Training data} & $n_d$ & \textbf{SR} & \textbf{kbps} \\
    \midrule
    A & 16k & LibriSpeech~\cite{librispeech} & 8 & 16k & 4 \\
    \midrule
    B1 & hifi\_16k\_320d & LibriTTS~\cite{zen19}, & 4 & 16k & 2 \\
    B2 & hifi\_16k\_320d\_large\_uni & VCTK~\cite{Veaux2016SUPERSEDEDC}, & 4 & 16k & 2 \\
    B3 & hifi\_24k\_320d & AISHELL~\cite{8384449} & 4 & 24k & 3 \\
    \midrule
    C & 24k\_320d & Valentini~\cite{valentini-botinhao2017noisy}  & 8 & 24k & 6.4 \\
    \midrule
    D & 24k & Same with E1$\sim$E5 & 32 & 24k & 24 \\
    \midrule
    E1 & 24k\_1.5bps & Common Voice~\cite{ardila-etal-2020-common},   & 2 & 24k & 1.5 \\
    E2 & 24k\_3bps & DNSC~\cite{dns},  & 4 & 24k & 3 \\
    E3 & 24k\_6bps &  Jamendo~\cite{bogdanov2019mtg}, & 8 & 24k & 6 \\
    E4 & 24k\_12bps & AudioSet~\cite{gemmeke2017audio},  & 16 & 24k & 12 \\
    E5 & 24k\_24bps & FSD50K~\cite{fonseca2021fsd50k}  & 32 & 24k & 24 \\
    \midrule
    F1 & en\_libritts\_16k\_gr1nq32ds320 &  & 32 & 16k & 16 \\
    F2 & en\_libritts\_16k\_gr8nq32ds320  & Subset of LibriTTS~\cite{zen19} & 32 & 16k & 16 \\
    F3 & en\_libritts\_16k\_nq32ds320 &  & 32 & 16k & 16 \\
    % \cline{3-3}
    F4 & zh\_en\_16k\_nq32ds320 & 25k hours collected data & 32 & 16k & 16 \\
    \bottomrule
\end{tabularx}
\vspace{-3mm}
\end{table}

This study investigates whether speech tokens obtained from various NAC models exhibit statistical properties similar to those of natural language. For the NAC models, we adopt six open-source codec models, where each of these models was trained under distinct conditions, including differences in datasets and bitrates, resulting in a total of 15 unique codec configurations shown in Table ~\ref{tab:nac_models} for comparative analysis~\cite{wu2024audiolanguagemodeling}. 

The extraction of token was conducted by the implementation of Codec-SUPERB~\cite{codecsuperb}.
To ensure consistency across models, we employed a standardised procedure for token sequence extraction and preprocessing. The speech signal was fed into a pretrained NAC model to extract a speech token sequence using a uniform codebook size of 1,024 for each dimension. Each token represented a 20~ms segment of input speech.
A deduplication process was also applied to eliminate consecutive repetitions of the same token, thereby preventing potential biases in the statistical analysis, particularly in $n$-gram distribution calculations.

Furthermore, to account for differences in dimension sizes $n_d$ caused by varying codec configurations across NAC models, we implemented an adjustment procedure. 
We treated each dimension of the NAC output as a separate token sequence and then concatenated them along with the time axis to create a single flattened token sequence. 
However, because each dimension uses an independent codebook but shares the same label space (e.g., 0 to 1,023), naive flattening would cause label collisions that obscure the actual distribution patterns. 
To resolve this, we offset each dimension’s token IDs by adding an index shift proportional to the dimension order (e.g., the first dimension uses IDs 0–1,023, the second uses 1,024–2,047, and so on). This effectively expands the overall codebook size while preserving the uniqueness of tokens across dimensions. Additionally, we prepared extra tokens that represented ``dimension start'' and ``dimension end'' and concatenated them at the beginning and the end of token sequences for each dimension, respectively. This process aims to mitigate undesired effects caused by artificial token overlap across codebooks during analysis.

The analysis was performed on 10 hours of single-speaker speech data for English (LJSpeech\footnote{\url{https://keithito.com/LJ-Speech-Dataset/}}) and Mandarin (Chinese CSS10~\cite{Park2019CSS10AC})
in order to investigate the language dependency of token. We used the manual transcriptions annotated to the corpora, and normalised through morphological analysis to obtain ground-truth (GT) word sequences for comparison.
\vspace{-1mm}

\section{Experimental Results}
\vspace{-1mm}
\subsection{Language-based Statistical Analyses}

For the speech token sequences obtained in this study, we generated $n$-gram sequences with $n = \{2, 3, 4, 6\}$ and conducted a series of statistical and structural analyses. Specifically, we evaluated the properties of the token sequences for each model on Zipf's law~\cite{zipf1949human}, Heaps' law~\cite{heaps1978information}, and calculations of entropy~\cite{shannon1951prediction} and redundancy. Each token sequences were compared by the natural language baseline, which is a word 1-gram.

\subsubsection{Token Frequency Distribution and Fit to Zipf's Law}

In this experiment, we computed the frequency distributions of token $n$-grams extracted from various NAC models and evaluated their fit to Zipf's law~\cite{zipf1949human} by analysing their rank-frequency distributions on a log-log scale.

For real-world data, when the tail is too heavy or too light, it is difficult to properly predict the shape of the data using general linear regression in Zipf's model~\cite{chan2024analyzinglanguagevisualtokens}. Thus, to robustly estimate the parameters of Zipf's law under such conditions, we employed the \texttt{powerlaw} library~\cite{clauset2009powerlaw}, which fits the rank-frequency data using the maximum likelihood estimate:
\[
\alpha = 1 + \frac{n}{\sum_{i=1}^{n} \log(x_i / x_{\min})},
\]
where \( x_i \) is the frequency of the \( i \)-th ranked token in the empirical distribution and \( x_{\min} \) works as a threshold, which is lower bound of the power-law behaviour in the rank-frequency distribution.
Here, \( \alpha = \eta + 1 \). Values of \( \alpha \) approaching 2 indicate a close adherence to Zipf’s law. Also, the goodness-of-fit was evaluated using the Kolmogorov–Smirnov (KS) distance, where larger suggest deviation from linearity~\cite{KSdistance}.

\begin{figure}
    \centering
    \includegraphics[width=\linewidth]{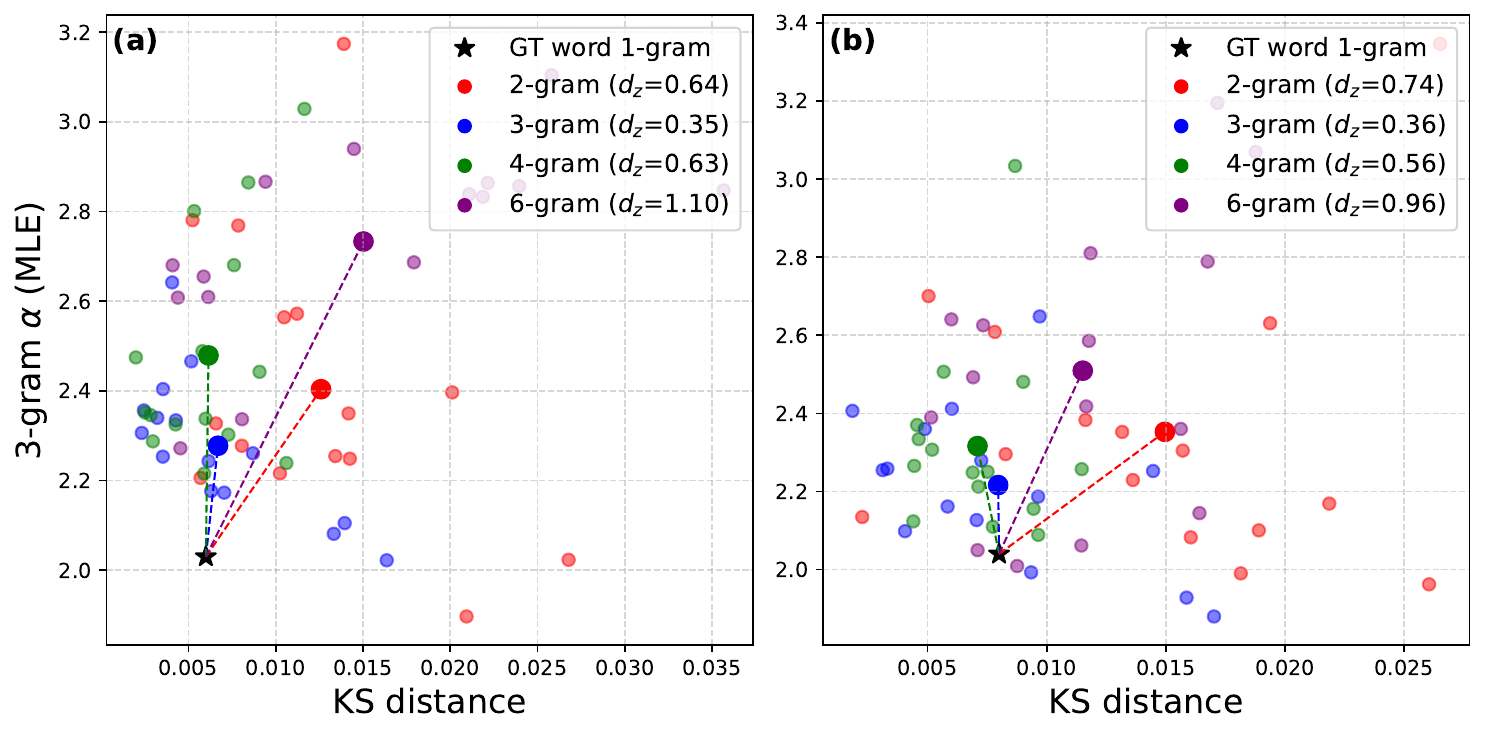}
    \caption{
    Relationship between KS distance and Zipf's law exponent ($\alpha$) for NAC token sequences compared to natural language reference (word 1-gram) for (a) English and (b) Chinese.
    }
    \label{fig:zipfs}
    \vspace{-3mm}
\end{figure}

\begin{figure}
    \centering
    \includegraphics[width=\linewidth]{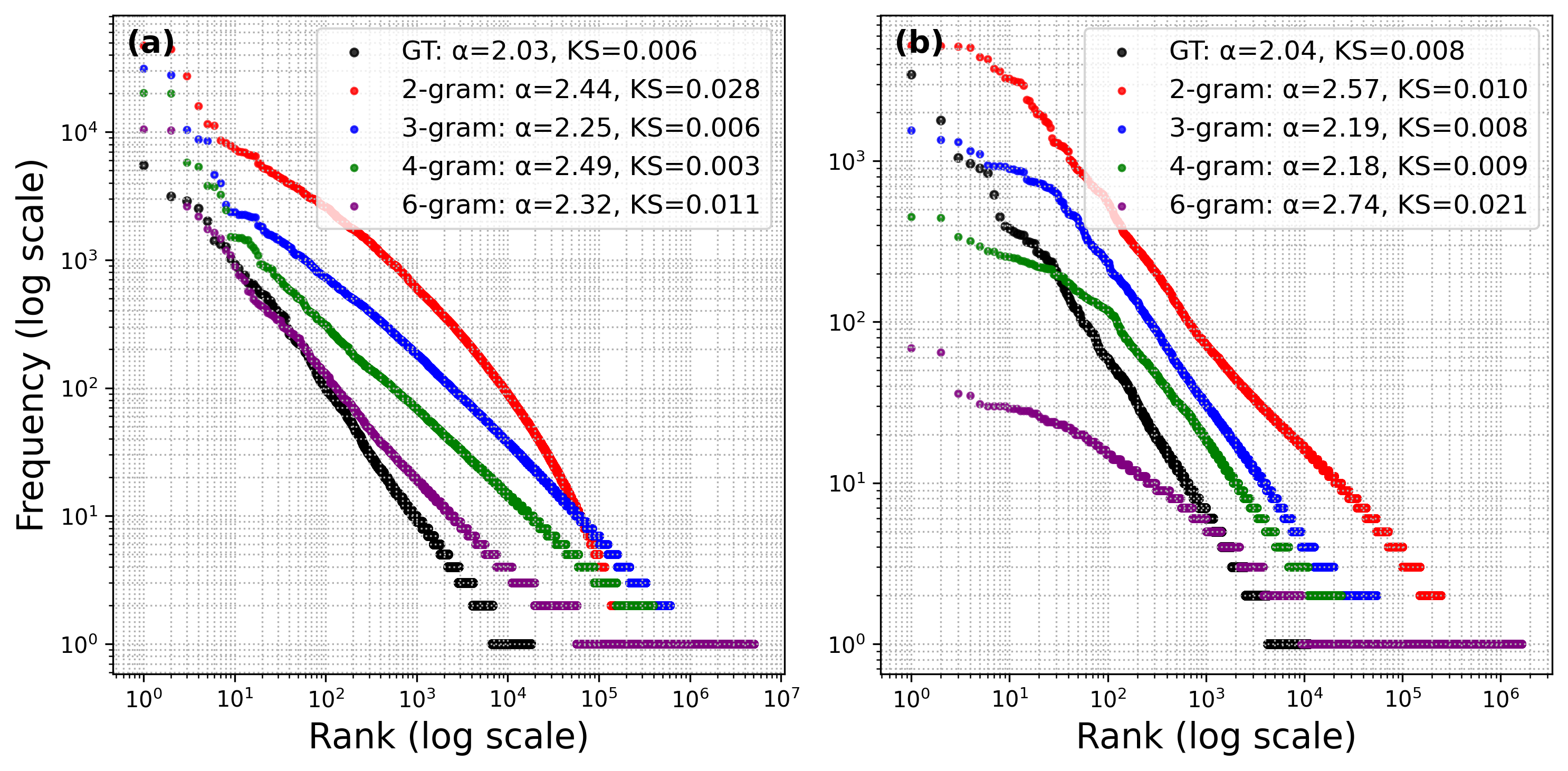}
    \caption{Comparison of $n$-gram distributions of the mean NAC token sequences with natural language reference (word 1-gram) for (a) English and (b) Chinese.}
    \label{fig:zipfs2}
    \vspace{-5mm}
\end{figure}

Figure~\ref{fig:zipfs} shows the results of the Zipf's law analysis. 
In this figure, each small dot represents one NAC model configuration, while the large dots represent the mean position of each $n$-gram group, calculated by averaging $\alpha$ and KS values. Distances from each $n$-gram mean to the GT were calculated using Z-score normalisation and displayed in the figure legend.
The token frequency and distribution including $\alpha$ and KS-distance values can be seen in Figure~\ref{fig:zipfs2}. For the mean NAC token, the entire token outputs from the model used in this study are stacked to represent the distribution of the entire token sequence in a single dimension, then divided by the total number of dimensions in the model, resulting in a normalized token with a codebook size of 1,024.

It shows that 3-gram token sequences consistently provided the best fit across the NAC models, demonstrating the most linear behaviour and thus the strongest alignment with the Zipfian distribution. This indicates that, at the 3-gram level, NAC tokens exhibit a frequency distribution where high-frequency tokens are used relatively frequently while low-frequency tokens remain present, albeit at lower frequencies. This pattern holds for both English and Chinese data, as showing the linear distribution graph in Figure~\ref{fig:zipfs2}.

\subsubsection{Token Novelty and Vocabulary Growth}

To evaluate the novelty of speech tokens, we conducted analyses based on Heaps' law~\cite{heaps1978information}. 
We analysed the natural language and the speech token sequences considering the token order, tracking the increase in unique tokens relative to the total number of tokens observed. 

\begin{figure}
    \centering
    \includegraphics[width=\linewidth]{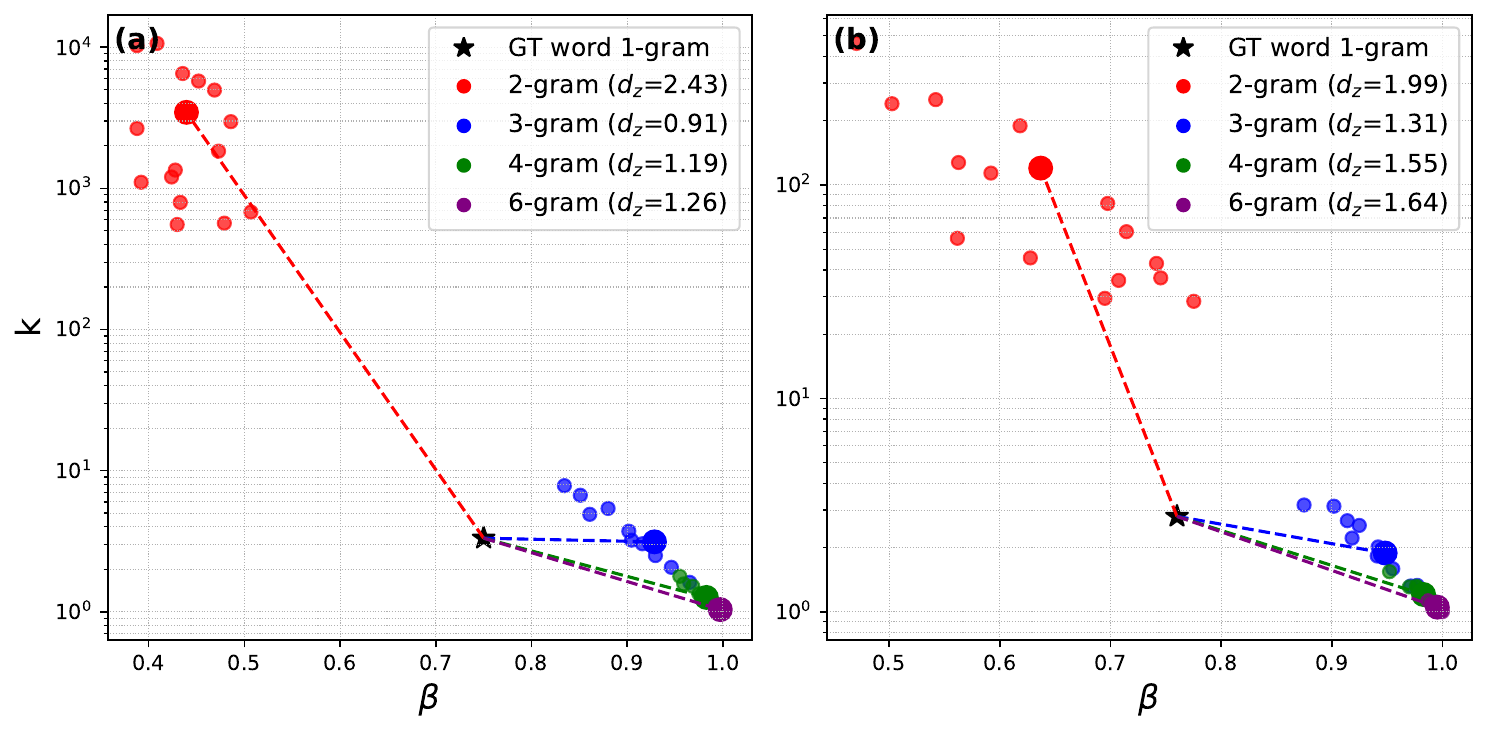}
    \caption{Relationship between Heap’s law components $k$ and $\beta$ for NAC token sequences compared to natural language reference (word 1-gram) for (a) English and (b) Chinese.}
    \label{fig:heaps}
    \vspace{-3mm}
\end{figure}

In Figure~\ref{fig:heaps}, we estimated both the scaling factor $k$ and the growth exponent $\beta$ for each model and $n$-gram setting. It is shown that 3-gram token sequences consistently display high similarities with those of natural language reference in both $k$ and $\beta$ values across both English and Chinese. 
Additionally, the variability of cluster assignment across different models was found to be substantial, indicating that some models tend to group tokens in ways that exhibit higher volatility, as reflected in the spread of $k$ and $\beta$ estimates.

\begin{figure}
    \centering
    \includegraphics[width=\linewidth]{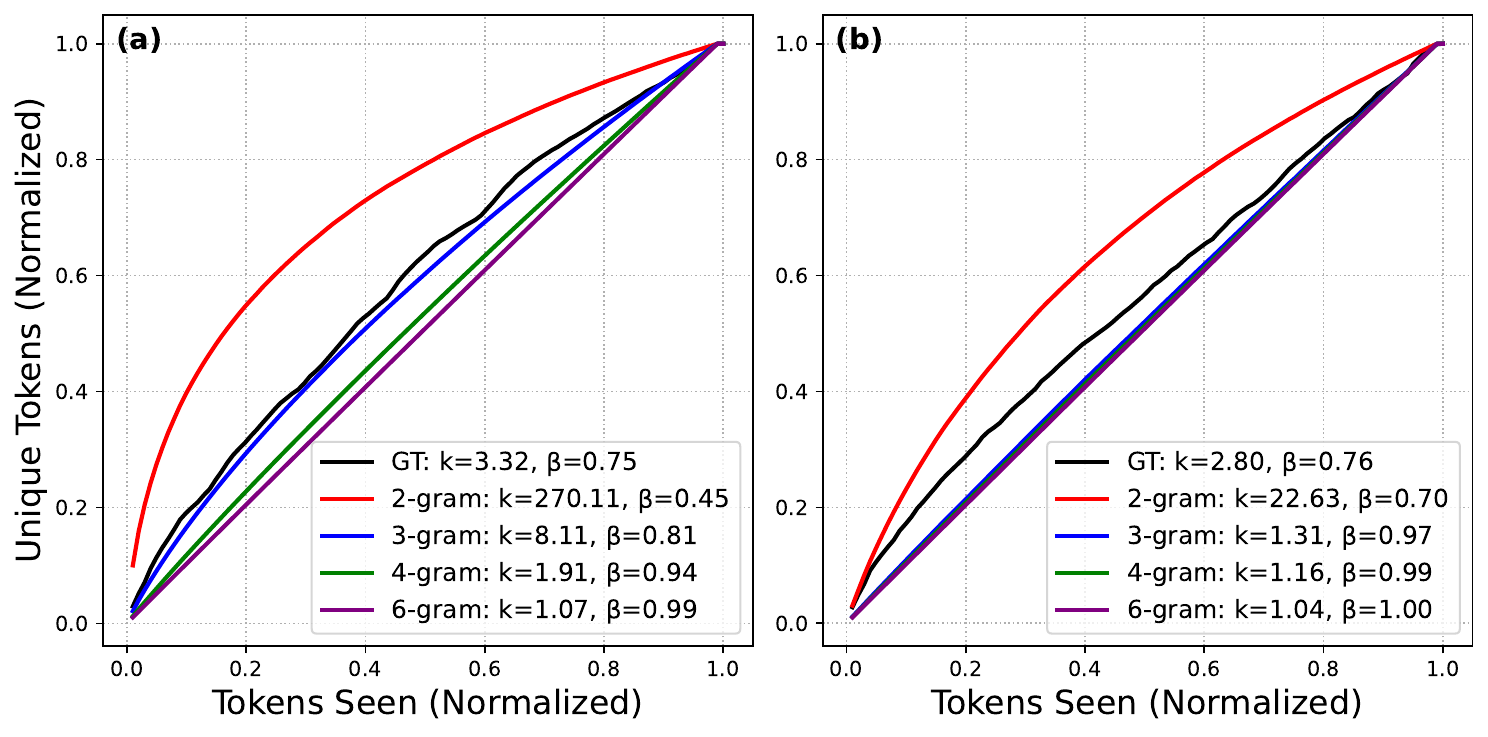}
    \caption{Comparison of $n$-gram token vocabulary growth of the mean NAC token sequences with natural language reference (word 1-gram) for (a) English and (b) Chinese.}
    \label{fig:heaps2}
    \vspace{-7mm}
\end{figure}

The results indicated that token sequences with higher $k$ values tended to exhibit sublinear growth, indicating repeated reuse of common tokens and reduced diversity, as shown in Figure~\ref{fig:heaps2}. Conversely, sequences with lower $k$ values displayed linear or near-linear growth, reflecting a more uniform introduction of new tokens. 
Furthermore, we observed that as $n$ increased, the complexity and diversity of the data also increased, leading to less repetitive patterns and larger vocabulary sizes. 

\subsubsection{Analysis of Token Entropy and Redundancy}

\begin{figure}[t]
    \centering
    \includegraphics[width=\linewidth]{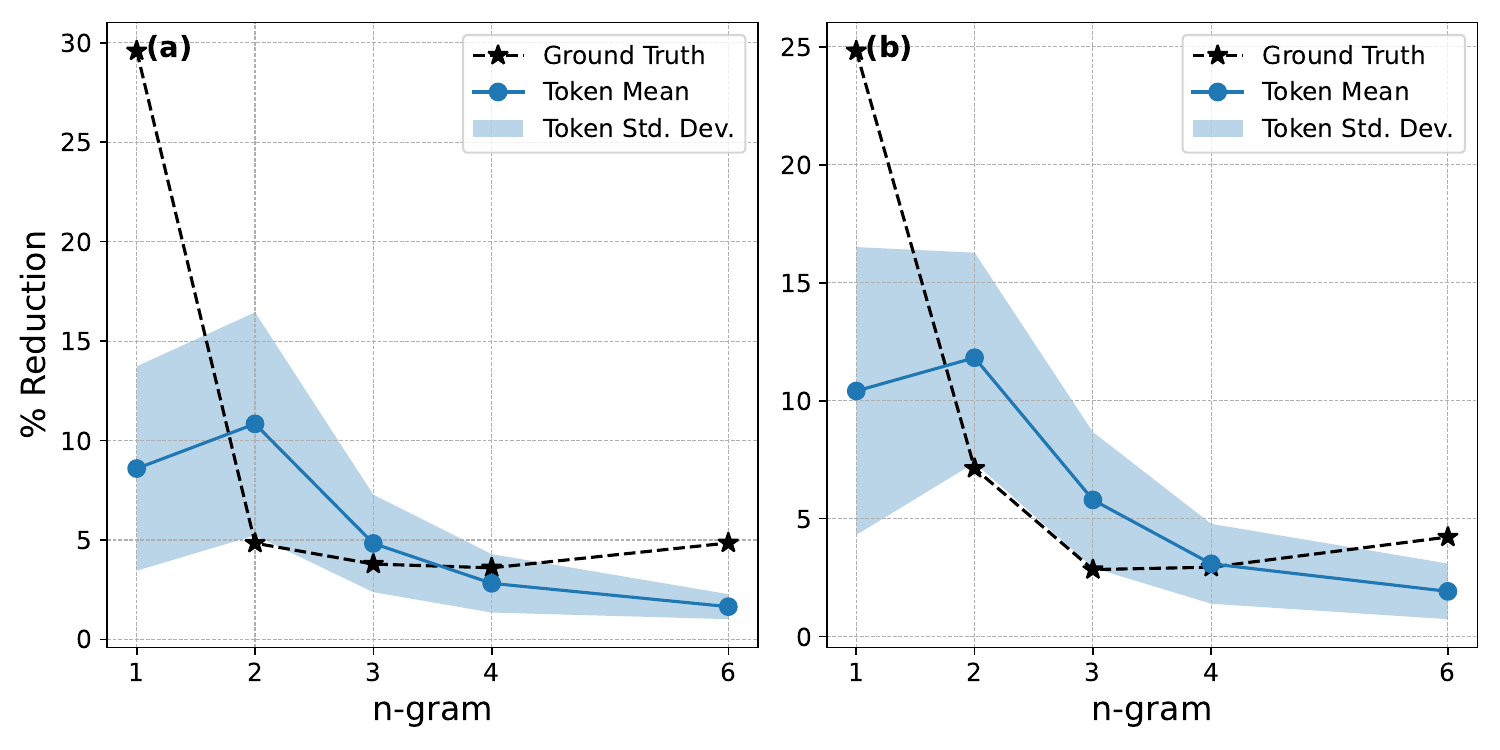}
    \caption{Comparison of token bit reduction rates across different $n$-gram configurations for NAC model tokens versus natural language reference (word $n$-grams) for (a) English and (b) Chinese. The results for 1-gram NAC tokens are also shown.}
    \label{fig:entropy}
    \vspace{-5mm}
\end{figure}

Finally, we evaluated the entropy and compressibility of token sequences generated by NAC models and compared them with natural language sequences. Figure~\ref{fig:entropy} shows the relationship between $n$-gram and the bit reduction rate. At the 1-gram level, natural language sequences exhibited a significantly higher bit reduction rate compared to NAC tokens. In contrast, NAC tokens displayed a relatively lower bit reduction rate at this level, suggesting a limited reuse of acoustic token patterns.

The bit reduction rate tend to drop sharply from 1-gram to 2-gram for natural language, while rather increased for token sequences. Interestingly, at the 2-gram and 3-gram levels, NAC token sequences showed higher bit reduction rates than natural language. This indicates that, at these intermediate $n$-gram levels, NAC token sequences contained acoustic patterns that frequently repeated, leading to increased redundancy that could be efficiently compressed using Huffman coding. However, at higher $n$-gram levels (4-gram and above), the bit reduction rate of NAC token sequences decreased sharply and fell below that of natural language. This suggests that NAC token sequences become highly sparse at these higher $n$-gram levels, as tokens no longer exhibit repetitive structures but rather fragment into near-unique combinations. 
These observations indicate that NAC token shows a high compression ratio in lower $n$-grams, but the effect decreases as $n$ becomes higher when modelling with more than 2-gram, in contrast to the natural language baseline. This difference indicates that NAC models may prioritize acoustic fidelity over statistical redundancy, resulting in less compressible token representations as $n$ gets bigger.

Also, it can be observed that between 3-gram and 4-gram configurations, natural language and tokens exhibit similar bit reduction rates. This suggests that the following $n$-gram distribution shows similar token distributions between natural language words and tokens.

\vspace{-1mm}
\subsection{Comparison with Speech Performance Benchmarks}
\vspace{-1mm}
This section investigates the relationship between the statistical properties of NAC tokens and speech performance benchmarks. We specifically examine whether tokens exhibiting more "language-like" statistics correlate with improved intelligibility (WER/CER) and naturalness (UTMOS).

To evaluate this relationship, we used the Codec-SUPERB~\cite{codecsuperb} implementation to decode the NAC tokens back into speech waveforms, generating resynthesised speech outputs for each codec model under study. 
For the intelligibility task, we then applied \texttt{whisper-medium}~\cite{radford2022whisper} ASR model, to transcribe the synthetic speech. By comparing these ASR outputs with the GT transcripts provided in the corpus, we calculated word error rate (WER) for English and character error rate (CER) for Chinese using the Levenshtein distance. This provided a quantitative measure of how closely the resynthesised speech aligns with the original linguistic content. 
Also, we assessed the naturalness of the synthetic speech using UTMOS~\cite{saeki22c_interspeech}, speech quality assessment metric that estimates the perceived naturalness of speech based on neural network architectures trained on large-scale human-annotated datasets. By applying UTMOS to the resynthesised speech samples, we were able to derive pseudo-naturalness scores that reflect how naturally the speech would be perceived by listeners. These benchmarks collectively allow us to analyse whether the statistical regularities observed in token sequences meaningfully impact the performance of speech processing systems in real-world tasks.

\subsubsection{Zipf’s Law}

\begin{figure}[t]
    \centering
    \includegraphics[width=\linewidth]{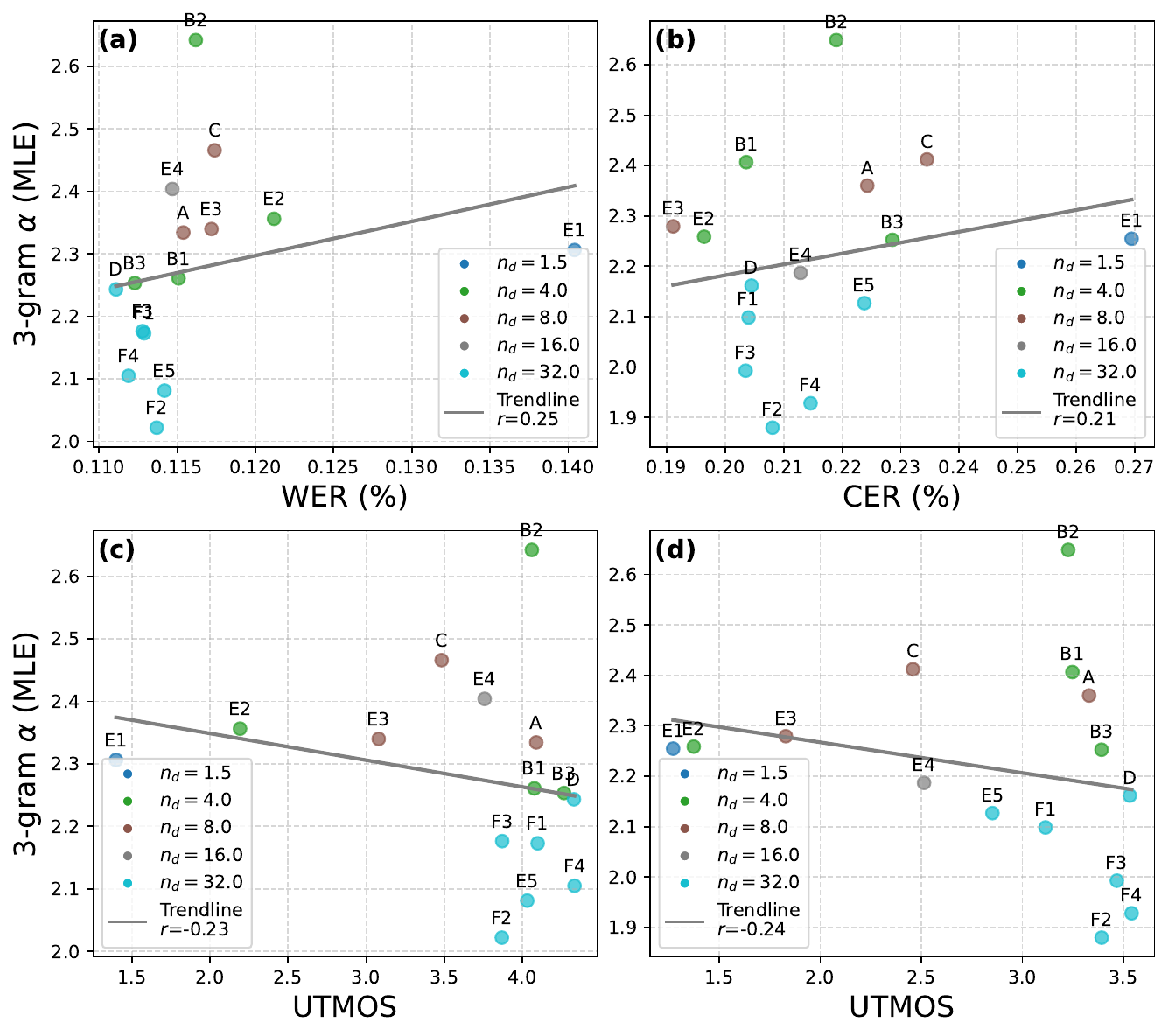}
    \caption{Relationship between the 3-gram Zipf’s law exponent ($\alpha$) and (a) English WER, (b) Chinese CER, (c) English and (d) Chinese UTMOS for speech token sequences. Each point represents each codec configurations, color-coded by dimension size $n_d$. The trendline was obtained by Pearson correlation coefficient $r$.}
    \label{fig:Zipf_asr}
    \vspace{-5mm}
\end{figure}

We utilized 3-gram token sequences against the benchmarks to investigate the relationship between token frequency and the performances in downstream tasks. 

Figure~\ref{fig:Zipf_asr} shows scatter plots of $\alpha$ values derived from 3-gram token sequences against the benchmark performances. The results reveal a consistent trend: as $\alpha$ decreases, where approaching the Zipf's law's theoretical value of 2, error rates tend to decrease while UTMOS scores increase. In addition, it is showed that model configurations with large dimensions displayed lower error rates and higher UTMOS, while also demonstrating lower $\alpha$. This suggests that token sequences exhibiting Zipfian behaviour with high-frequency token usage correlate with improved speech recognition accuracy and perceived naturalness. 

Interestingly, we observed that this trend was most clearly observed with 3-gram token sequences, while analyses with 2-gram and 4-gram configurations did not show the same consistent relationship between $\alpha$ and benchmark performances.\footnote{\label{note1}The exact values and graphs could not be included in the paper due to page limitations.} This observation suggests that 3-gram sequences may strike an optimal balance between capturing local linguistic patterns and maintaining token diversity, which in turn aligns well with both statistical regularity and acoustic information preservation. In contrast, 2-grams may be too short to adequately represent contextual dependencies, while 4-grams may introduce excessive sparsity and volatility in token usage.

\subsubsection{Heaps' Law}

\begin{figure}[t]
    \centering
    \includegraphics[width=\linewidth]{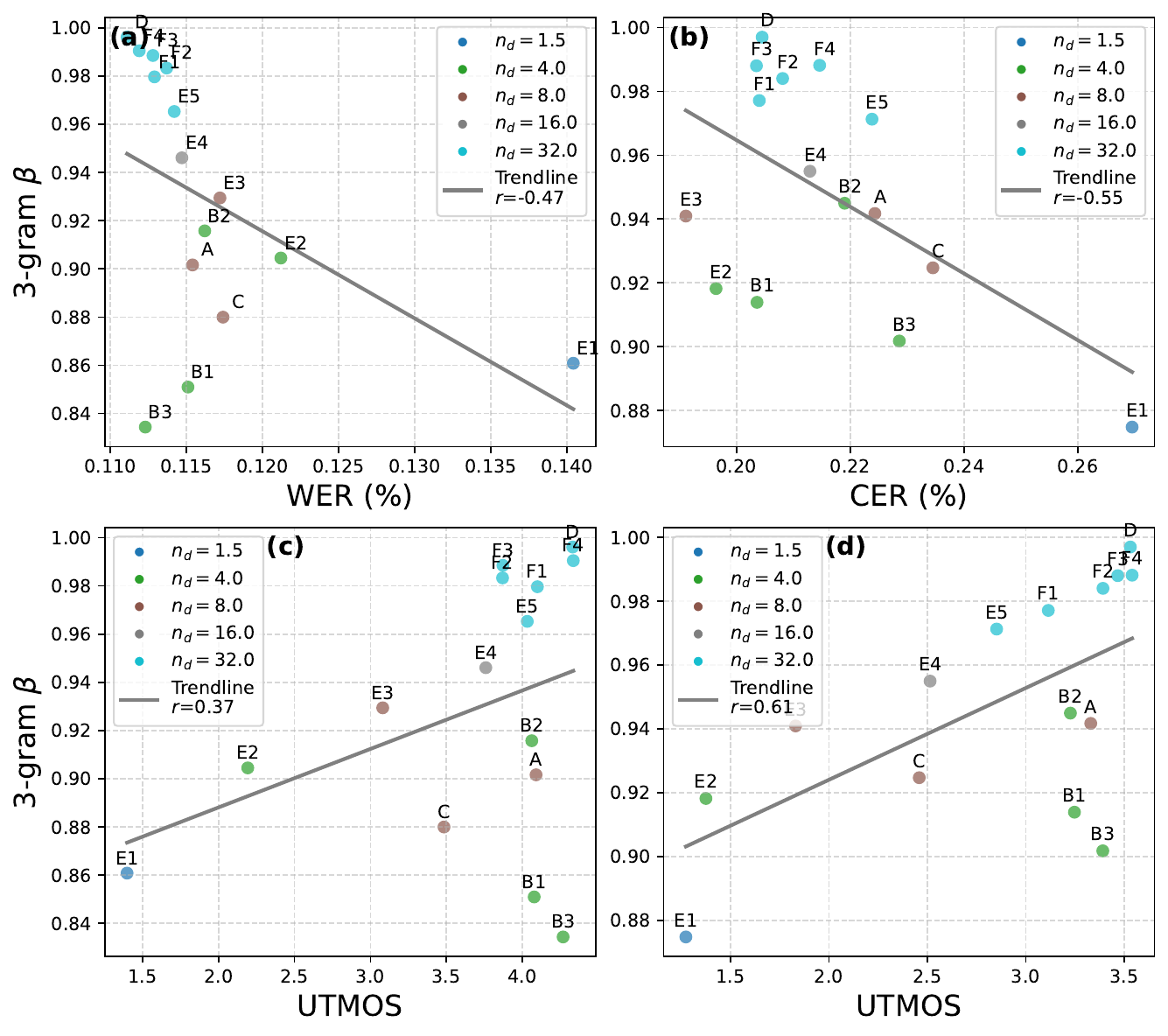}
    \caption{Relationship between the 3-gram Heap’s law exponent ($\beta$) and (a) English WER, (b) Chinese CER, (c) English and (d) Chinese UTMOS for speech token sequences. Each point represents each codec configurations, color-coded by dimension size $n_d$. The trendline was obtained by Pearson correlation coefficient $r$.}
    \label{fig:heap_asr}
    \vspace{-5mm}
\end{figure}

We analysed the correspondence between the Heaps' law exponent $\beta$ and $k$—estimated from 3-gram token sequences—and speech benchmarks. Figure~\ref{fig:heaps} shows scatter plots of $\beta$ values against the benchmark performances. The results reveal a consistent trend: as $\beta$ approaches 1 (indicating a more linear growth of the vocabulary size with respect to the number of observed tokens), error rates (WER and CER) decrease, and UTMOS scores increase. It is also showed that model with a configurations of large dimensions displayed higher $\beta$. This finding suggests that token sequences exhibiting near-linear vocabulary growth—indicative of consistent token usage—correlate with improved speech recognition accuracy and perceived naturalness. This trend was robust across different $n$-gram configurations, with 2-grams to 6-grams consistently demonstrating that higher $\beta$ values align with better benchmark performances, except for a few outliers. This reinforces the conclusion that token sequences capturing stable vocabulary expansion contribute positively to both recognition and synthesis.

Additionally, the scaling factor $k$ in the Heaps' law formulation was also found to play an important role\footnotemark[\value{footnote}]. As $k$ approached 1, meaning that the model effectively captures each token's presence without excessive redundancy, error rates decreased and UTMOS scores improved. 
These results collectively highlight that tokens adhering to linguistic patterns consistent with natural language. 

\subsubsection{Entropy}

\begin{figure}[t]
    \centering
    \includegraphics[width=\linewidth]{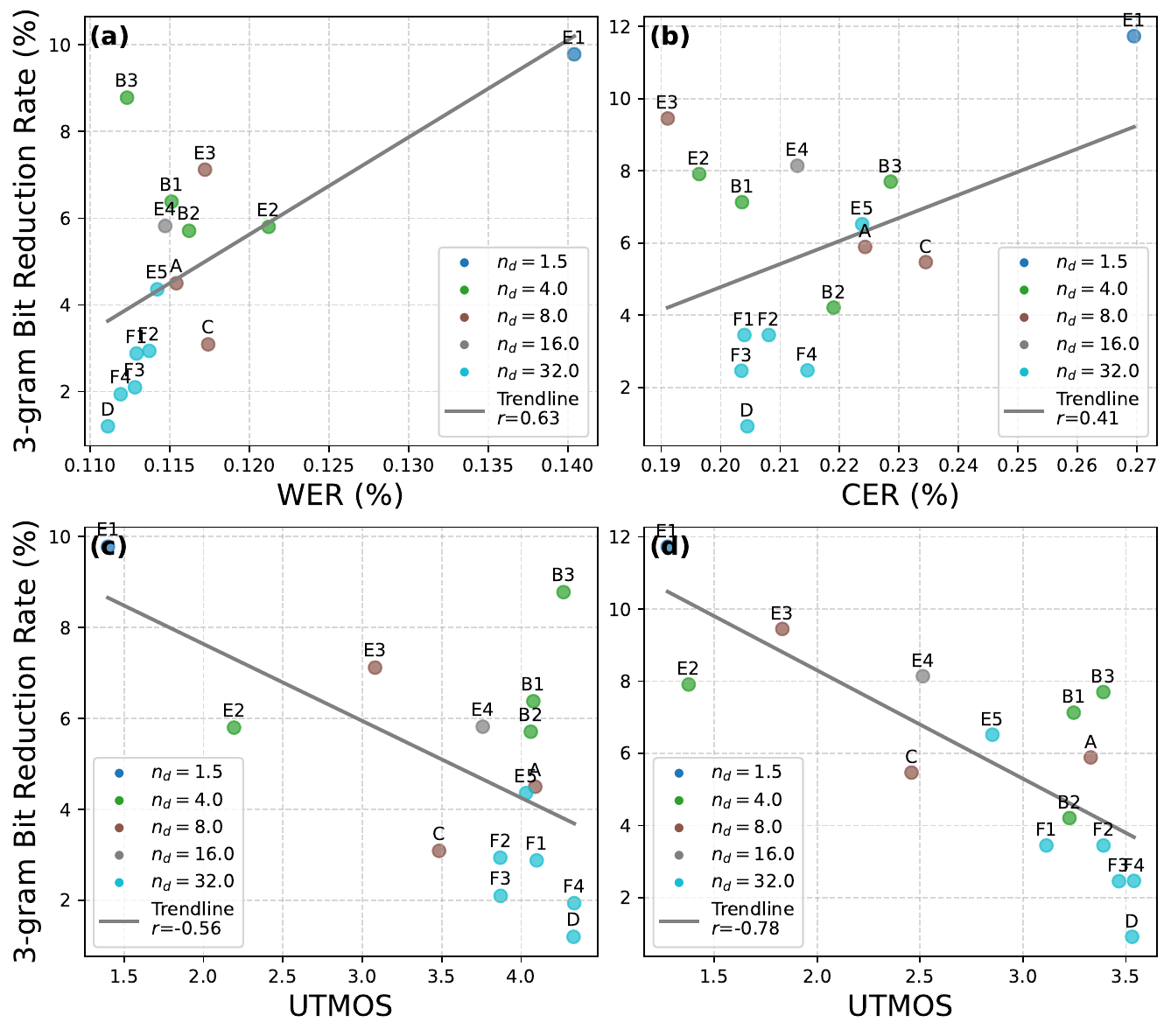}
    \caption{Relationship between the 3-gram token bit reduction rate and (a) English WER, (b) Chinese CER, (c) English and (d) Chinese UTMOS for speech token sequences. Each point represents each codec configurations, color-coded by dimension size $n_d$. The trendline was obtained by Pearson correlation coefficient.}
    \label{fig:entropy_asr}
    \vspace{-5mm}
\end{figure}

We analysed the bit reduction rate derived from 3-gram token sequences and its relationship with speech quality benchmarks. Figure~\ref{fig:entropy_asr} presents scatter plots showing these relationships. The analysis reveals a consistent trend: as the bit reduction rate approaches zero—indicating that the token sequences are less compressible—error rates decrease and UTMOS scores increase. It is also showed that model with a configurations of large dimensions displayed lower bit reduction rate. This suggests that token sequences with lower redundancy and higher diversity are more closely aligned with the underlying structure of natural language and correspondingly support better speech quality. This pattern was also consistently observed across $n$-gram configurations from 2-grams to 6-grams\footnotemark[\value{footnote}], indicating that lower bit reduction rate across $n$-gram reliably correlate with improved recognition accuracy and higher perceived naturalness. 

\vspace{-2mm}
\section{Conclusion}

In this study, we explored in what cases speech token sequences from NAC models exhibit statistical properties similar to those observed in natural language. Overall, NAC tokens consistently aligned closely with natural language distributions at the 3-gram level, and trends in the Zipf's law exponent ($\alpha$), Heaps' law parameters ($\beta$ and $k$), and bit reduction rate all indicated that token sequences with higher linguistic regularity and lower redundancy contributed to improved speech representation quality. Together, these results indicate that certain linguistic statistical properties of NAC tokens, particularly at specific $n$-gram granularities, play an important role in bridging the gap between acoustic fidelity and linguistic structure in speech processing. 

Beyond the contributions here, we believe that it is insightful to further explore the case of real-world, multi-speaker audio in noisy environments, and NACs designed for multilingual domains. We hope such explorations can yield further insights into how we can design token representations that balance linguistic regularity and acoustic detail, supporting high-quality speech generation and accurate language understanding.

\textbf{Acknowledgment:}
The work was supported by JST Moonshot Grant Number JPMJMS2011 and based on results obtained from a project outsourced by the New Energy and Industrial Technology Development Organization (NEDO).

\bibliographystyle{IEEEtran}
\bibliography{bib/tts}

\end{document}